\documentclass[acmtog]{acmart}
\renewcommand\footnotetextcopyrightpermission[1]{} % removes footnote with conference information in first column
\AtBeginDocument{%
  \providecommand\BibTeX{{%
    \normalfont B\kern-0.5em{\scshape i\kern-0.25em b}\kern-0.8em\TeX}}}

\setcopyright{acmcopyright}
\copyrightyear{2019}
\usepackage{amsmath} % assumes amsmath package installed
\usepackage{amssymb}  % assumes amsmath package installed
\usepackage{algorithm2e}
\usepackage{graphicx}
\usepackage{subcaption}
\usepackage{caption}
\usepackage{multirow}
\usepackage{multicol}
\usepackage{slashbox}
\begin{document}

\title{Robot navigation and target capturing using nature-inspired approaches in a dynamic environment}

\author{Devansh Verma}
\authornote{Both authors contributed equally to this research.}
\email{devanshverma951@gmail.com}
\author{Priyansh Saxena}
\authornotemark[1]
\email{saxenapriyanshasd@gmail.com}
\orcid{0000-0003-1407-9752}
\affiliation{%
  \institution{ABV-Indian Institute of Information Technology and Management, Gwalior}
  \city{Gwalior}
  \state{Madhya Pradesh}
  \country{IN}
  \postcode{474015}
}

\author{Ritu Tiwari}
\affiliation{%
  \institution{Indian Institute of Information Technology, Pune}
\city{Pune}
  \state{Maharashtra}
  \country{IN}}
\email{tiwariritu2@gmail.com}
\renewcommand{\shortauthors}{Devansh and Priyansh, et al.}

\begin{abstract}
  Path Planning and target searching in a three-dimensional environment is a challenging task in the field of robotics. It is an optimization problem as the path from source to destination has to be optimal. This paper aims to generate a collision-free trajectory in a dynamic environment. The path planning problem has sought to be of extreme importance in the military, search and rescue missions and in life-saving tasks. During its operation, the unmanned air vehicle operates in a hostile environment, and faster replanning is needed to reach the target as optimally as possible. This paper presents a novel approach of hierarchical planning using multiresolution abstract levels for faster replanning. Economic constraints like path length, total path planning time and the number of turns are taken into consideration that mandate the use of cost functions. Experimental results show that the hierarchical version of GSO gives better performance compared to the BBO, IWO and their hierarchical versions.
\end{abstract}


\maketitle

\section{Introduction}\label{Inroduction}
The modern-day robots are made adept in performing several complex tasks by themselves, which usually required human assistance. Some of the recent developments like the Mars Spirit Robot and Self-Driving Cars \cite{levinson2011towards} are an excellent example of today's technological advancements in the fields of mobile robots. The robots of today are extremely intelligent and are capable of finding their way of completing the task given to them. One of the most crucial functions of mobile robots is to move from one place to another safely without damaging themselves or their corresponding environment.

Robotic navigation is one of the toughest robotic tasks. This problem can be divided into different categories by the following criteria:

\begin{itemize}
    \item \textbf{Type of Environment}:
    The type of environment can be broadly classified as 2D and 3D \cite{pandey2018three}. The algorithms employed for both of them vary in terms of complexity, approach and applicability. The 2D environment involves two coordinates with the third coordinate is constant, whereas in 3D, the third coordinate is also varying.\\
    
    \item \textbf{Nature of Environment}:
   It refers to the environment in which the robot operates, which can be either static or dynamic. Static environment refers to an environment in which the number and the position of obstacles remain constant with respect to time \cite{masehian2006tabu}. Due to static nature obstacles, global path planning algorithms can be applied efficiently in order to solve the problem. On the contrary, in a dynamic environment, the environment can change while an agent is deliberating. The speed, position as well as the shape of an obstacle may change with time. Since the same area needs to be explored, again and again, global path planning cannot be applied, so the only choice is local path planning.\\
    
    \item \textbf{Planning approach}:
    The robot navigation problem can be classified based on knowledge gained at a local and global level. In the case of global path planning, the agent has full prior knowledge about shape, orientation, location and arrangement of the obstacles. Using this information, the path planning algorithm minimizes the cost function in order to reach the target. This approach can only be applied in case of a static environment where obstacles do not move. On the contrary, in local path planning, the robot has little to no prior information about the environment and surroundings. The robot senses the environment within its sensory range and with the help of some meta-heuristic approach plans its next steps while moving in the environment.\\
\end{itemize}

\par Target Searching is a problem of estimation of the location of the target(s) in a given region. This region can be air, water or land. The targets can be mobile as well as stationary and can be of evasive or non-evasive nature \cite{albert2018survey}. The robot only has to search for the specific area only once in the case of static targets. On the contrary, for dynamic targets, the same area needs to be explored multiple times as the probability of finding the target is non-zero in that area since the area can be occupied again. For simulation, the target is assumed to be a point object. In practical scenarios, the target can be a person, an animal, wildfire or even a flying aircraft. There are many application such as in search and rescue operations in disastrous areas \cite{kantor2003distributed}, robots that can be used for locating mines \cite{acar2003path}, robots used for planetary exploration \cite{schilling1996mobile}, and in search of remote fires \cite{marjovi2009multi}.\\

\par Nature has always been a constant source of inspiration for humans \cite{shukla2017discrete}. Natural phenomenons have motivated us immensely and enabled us for better innovation. Many laws and theories have been formulated observing different aspects of nature, whether it be plants, animals, birds or whatnot. Nature Inspired Algorithms are algorithms inspired by natural processes regarding biological, geographical, and chemical processes that take place in the natural world. These algorithms have immense potential for solving many real-life problems. A tremendous amount of research has taken place in the past two decades in this field.
 \\

\section{Related Works}\label{rel-works}
\par The modern-day robotics is pro efficient in performing several complex tasks in intelligent and automated fashion \cite{pandey2018three}. It is also one of the most difficult as well as a challenging task in a multi-robot system. The implementation of real-life constraints like energy consumption travel cost, path length also plays a significant role in complicating the task even further \cite{janis2016path}. 

The path planning can be done both in the air as well as in water \cite{meghjani2016multi} and also in 2D or 3D environment. Several Machine Learning and nature-inspired algorithms like A*, Bat Algorithm \cite{wang2016three}, PSO \cite{bilbeisi2015pso} are applied too. Many existing algorithms, like Ant Colony Algorithm \cite{zhang2017improved}, Invasive Weed Optimization \cite{mohanty2014new}, have been applied in order to tackle this problem. Path planning is purely an optimization problem as it requires optimizing the cost from source to the destination. It is an NP-hard problem and therefore, there is a need for a framework for autonomous robots \cite{liu2018autorobot} to optimally find their paths. In \cite{phillips2011sipp}, the problem has been solved by dividing time into the safe and unsafe interval and acting accordingly. The problem has also been solved using the fuzzy logic approach \cite{pandey2014path}. In \cite{drake2018mobile} incremental path planning algorithm for dynamic target has been used that uses past data to update path by sacrificing optimally for reducing computational complexity, and in \cite{julia2012searching} estimation is used in order to search for the target.One aim of any path planning problem is reaching the target. This target can be anything from a big object to a small object. The prime objective of the navigation problem is aimed at seizing the target. The target can be fixed or mobile. In case of a fixed target, the robot only needs to look for the specific area once. On the contrary, for moving target, the same area needs to be explored multiple times as the probability of finding the target is non-zero at that point as the area can be occupied again. The targets can be located anywhere, be it land or water \cite{meghjani2016multi}. For detection of a target, specific Gaussian Mixture Models are used in cyber cars \cite{premebida2006multi}. The target may also live for some time and may disappear after that \cite{jain2017hybridization}. In \cite{zhang2018multi}, target tracking is done based on some patterns. The target searching problem can be done with or without \cite{chaudhary2012detecting} previous information.

In \cite{dadgar2016pso} PSO is used with multiple robots for pathfinding. In \cite{jensen2018online} there has been a comparison between various algorithms for multi-robot discipline. Deep CNN has also been applied in order to solve this problem \cite{cui2017end}. In  \cite{saldana2015distributed} coordination among the robots is done for detection and tracking of anomalies. In\cite{hayat2015mobile} SA algorithm is used to achieve an optimum collision-free path with mobile robots in a static environment containing circular obstacles. The creation of the map is not required in order to solve the path planning problem \cite{premebida2006multi}. In \cite{zhang2018multi}, multiple robots coordinate among them self treating the densities of robots and targets as properties of the environment.

\subsection{Research Gaps}
Although much research has been done in the field of robot navigation and targets capturing, it is still not sufficient enough. This paper tries to overcome some of these research gaps through the proposed techniques. Here are some of the research gaps present in this field of robot navigation and target capturing:

\begin{enumerate}
    \item The solution obtained by the deterministic algorithm is unfeasible as it is time-consuming. The meta-heuristic algorithms provide sub-optimal paths but in very less time as compared to that of deterministic algorithms.
    
    \item Not much work has been done in the field of multiresolution path planning using abstract levels. The use of hierarchical levels results in faster replanning.
    
%     \item Not much emphasis has been given on multi robot path planning and exploration
% using nature inspired algorithm
\end{enumerate}

\section{Methodology}\label{methodology}
\subsubsection{Proposed Workflow}
The complete project is divided into following modules: 

\begin{itemize}
    \item \textbf{Environment Creation}: The first step is to create an arena for simulation. It includes designing and modelling the static and dynamic target and obstacles.
    
    % \item \textbf{Cost function}:
    
    \item \textbf{Algorithm Selection}: TIn the next step, the objective is to devise a strategy to generate collision-free paths which involve continuous sensing of the environment and frequent replanning.
    
    \item \textbf{Path smoothening and refinement}: The trajectory generated by the path planning algorithms is generally not smooth and contains several loops and sharp turns. Path smoothening methods such as \textit{Catmull- Rom splines} are used to generate smooth, realistic paths.
\end{itemize}

\par
\begin{figure*}
\centering
\includegraphics[scale=0.28]{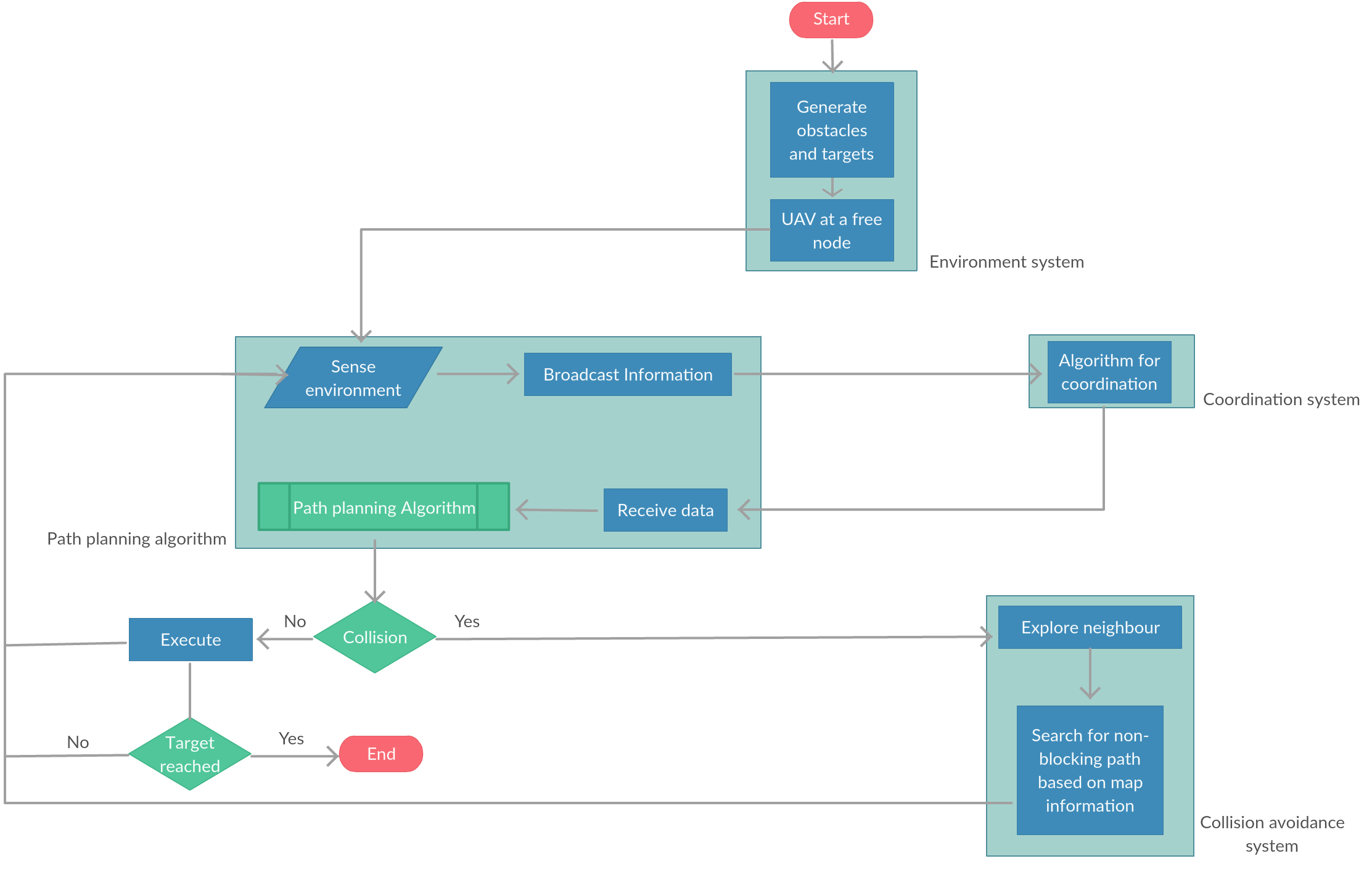}
% \centerline{\includegraphics[width=\textwidth,height=15cm]{images/Workflow.png}}
\caption{Proposed methodology}
\label{fig:1}
\end{figure*}

 The proposed methodology is shown in Figure \ref{fig:1}. It starts with the creation of the arena, which includes adding obstacles and targets. The obstacles and targets are both static as well as dynamic and time-varying. In the path planning phase, the robot senses the environment and broadcast this information to the semi-centralized server, which is employed to achieve the coordination task among the robots in case of a multi-robot system. The received data from the server is used for path planning and moving towards the target. If there is no collision detected, the robot is moved to the required position and checked for target reachability. If a possible collision is detected, the collision avoidance algorithm is triggered, which includes different strategies for dealing with static and dynamic obstacles. After collision avoidance, the path planning system is kicked again, and the process continues.

\subsubsection{Glowworm Swarm Optimization(GSO)}
It is a swarm intelligence based algorithm inspired by a lightning worm called a glowworm. It was proposed by \textit{K.N. Krishnand and D. Ghose} in the year 2005. GSO has a wide variety of applications like localization, swarm intelligence, clustering, routing and many more.

Glowworm is a small worm that emits light due to a light-emitting compound called luciferin. Each glowworm is capable of adjusting its luciferin content based on environmental factors. The light thus generated is used to lure the prey or attract other worms. It is this behaviour of glowworm is exploited in the GSO algorithm for solving various optimization problems. The worms in the swarm are distributed randomly in search space, having some initial luciferin value. The higher level of luciferin content means a better position of glowworm in the search space. The glowworms then calculate the fitness in the current space using some fitness function and then broadcast this information to all the neighbours within the search radius. The glowworm is attracted to other worm having higher luciferin content and within its search space. These movements which are derived based on the local information provided by the neighbours and selective interaction with the neighbours enables the swarm to divided into disjoint clusters which converges at different points containing the local optima to find multiple solutions. It is similar to other swarming algorithms like ACO and PSO.

The GSO algorithm works in several phases which are responsible for the movement of glowworm in the search space. Figure 2 shows a flowchart of the GSO.  It starts by adding a randomly generated population of n glowworms. All the glowworms have same initial luciferin content. The algorithm starts with luciferin-updation phase followed by the movement phase. The phases in GSO can be explained as follows: - \\

\textbf{Glow-worm Distribution Phase:}
This is the first step of the algorithm. The swarm of glowworm is randomly dispersed in the search space each glowworm having an initial luciferin value of $l_0$ and a search radius $r_0$. \\

\textbf{Luciferin Updatation Phase:} 
This phase changes the luciferin content of the glowworms based on the function value at the position of the glowworm. The luciferin content is enhanced based on the objective function value. Simultaneously the value is decreased to simulate the degradation of luciferin with time. The luciferin update phase can be stated as follows:\\ 
		\begin{equation}
		    l_{i}(t)=(1-\rho) l_i (t-1)+\gamma J(x_i (t))
		\end{equation}
 where $\rho$ and $\gamma$ are decay constant and luciferin enhancement constants respectively.$l_{i}(t)$ denotes the luciferin content of $i^{th}$ glowworm at $t^{th}$ iteration. $J(x_{i}(t))$ denotes the objective function. \\
 
 \textbf{Movement Phase:}
 In this phase the glowworm decides to move towards other neighbouring glowworms based on probabilistic mechanism. The glowworm is attracted to other glowworm which is within its range and have luciferin content greater than its own. The equation governing the probability to move towards neighbour glowworm is as follows:\\
	\begin{equation}
	    p_{ij} (t)=\frac{l_j (t)-l_i (t)}{\sum\nolimits_{k \in N_{t}}(l_k (t)-l_i (t)}
	\end{equation}
where $j$ is an ordered set of neighbouring glow-wworm, according to the below formula:\\
    \begin{equation}
	   j\in N_i (t), N_i (t)={j:d_{ij} (t)<r_d^i (t);l_i (t)<l_j (t)}
	\end{equation}
where $d_{ij}$ denotes the distance between $i^{th}$ and $j^{th}$ glowworm and the search radius of $i^{th}$ glowworm is denoted by $r_{d}^{i}$. After that, the movement of glowworm is governed by the following equation:\\

\begin{equation}
    x_i (t+1)=x_i (t)+s \times \frac{(x_j (t)-x_i (t))}{\parallel x_j (t)-x_i (t)\parallel }
\end{equation}
where $x_{i}$(t+1) denotes the position of $i^{th}$ glowworm at t+1 th iteration, s shows the step size and $\parallel.\parallel$ denotes the euclidean distance operator. which takes the luciferin content of the glowworms.\\

\textbf{Neighborhood Range Update Phase:}
Each glowworm is assigned a range which changes with each iteration so that it doesn't fall for local optimum and can efficiently reach the global optimum. Therefore GSO uses adaptive neighbourhood range to detect multiple peaks of the function. The updation of range equation is shown as:

\begin{equation}
r_d^i (t+1)=min \{ r_s,max \{ 0,r_d^i (t)+\beta (n_t-N_i (t)) \}\}
\end{equation}

where $r_d^i (t+1)$ denotes the dynamic range of i th glowworm at t+1 th iteration, $r_{s}$ denotes the minimum sensory radius, $\beta$ is a constant and $n(t)$ is a parameter for controlling the neighbourhood.\\

\begin{figure}[hbt!]
	\begin{center}
		\includegraphics[scale=0.65]{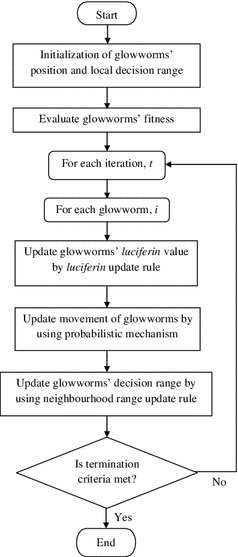}
		\caption{Glowworm Swarm Optimization Algorithm}
	\end{center}
\end{figure}

\subsubsection{Invasive Weed Optimization (IWO)}
This is also a swarm intelligence based algorithm. It is inspired by the colonizing property of the weeds. It was initially presented by \textit{Mehrabian and Lucas} in 2006. The primary motivation of this algorithm is from the advance and development characteristics of the weed plants that grow in the field. These plants don't require many resources and exploit the available resources to grow and reproduce. This is the reason for their existence to date, even if humans want to tear them off the field. Weeds have proved very vigorous and adaptable to environmental changes. They have a high survival probability and capable of adaptation.\\ 

In the modern agricultural system, many resources are left in the field, creating a budding spot for the weed plant. Weeds take advantage of this and invade the field suing these resources. They colonize, reproduce and disperse new seeds based on their adaptability to nature. The weeds having higher fitness as compared to conventional crops are capable of generating a large number of seeds. These three steps continue until all field reserves are depleted, leaving behind the weeds with high adaptability. This hard competition amongst the plants force them to get adapted with the environment making them more robust to the changes.

 The IWO algorithm is generally used to solve several optimization issues using the fashion mentioned above. Compared with some algorithms, it has some special features like colonize, disperse, reproduce and competitive exclusion. Figure 3 shows a flowchart of the IWO. Following are the phases to implement IWO:\\

\begin{itemize}
\item{\bf Initialization: } \\ Various parameters like population size, number of iterations, constants related to IWO in the search space.

\item{\bf Initializing a population: } \\ All the weeds present initially in the solution space generate an initializing path to the destination from the start. This solution is enhanced in the further iterations.

\item{\bf Reproduction: } \\Once an initial solution has been formulated, each one is enhanced. Based on the fitness values of each weed, it generates seeds. Compared to less fitting weeds, fitter plants generate more seed .There is a minimum and maximum cap on number of seeds produced per weed plant.

\item{\bf Spatial dispersal: } \\ Once the seeds are created, they are spread to the nearby position around the parent. This ensures that more seed is produced around the parent. Below equation represents the change of variance with respect to the iterations.\\
\small
{
\begin{eqnarray}
\sigma_{iter}=\left(\frac{iter_{max}-iter}{iter_{max}}\right)^{n} \times (\sigma_{initial}-\sigma_{final})+\sigma_{final}
\end{eqnarray}
}
where, $iter_{max}$ and iter depict the maximum number of iterations and current iteration. The $\sigma_{final}$ and $\sigma_{initial}$ are the final and initial standard deviations and n is a nonlinear modulation index.\\

\item{\bf Competitive exclusion: } \\Once all the process  is completed, the weeds are sorted by the fitness value. The total number of weeds exceed the population size, the further population is discarded  from the following iterations. The fitter plants produce more seeds to contribute there role in the solution. This instrument utilizes the standard of survival of the fittest strategy.
\end{itemize}
\vspace{0.2in}

\begin{figure}[h!]
\centerline{\includegraphics[width=3.5in]{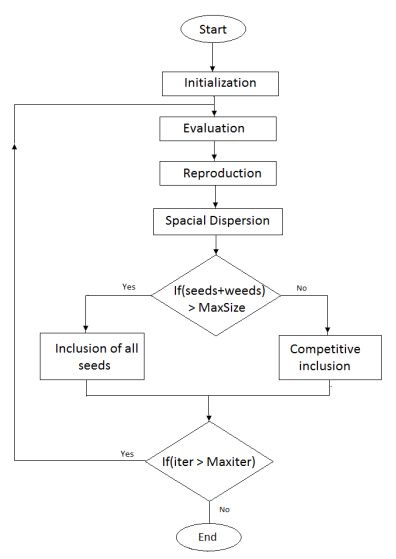}}
\caption{Invasive Weed Optimization Algorithm}
\end{figure}

\subsubsection{Biogeography-Based Optimization (BBO)}
This technique was formulated in 2008 by \textit{Dan Simon} \cite{simon2008biogeography}. The above-stated technique is based on the principles of bio-geography. It is the study of the geographical distribution of biological organisms.

\subsubsection{Cost Function}
Cost function is a function of various factors like length, velocity, height etc. which is used to govern the direction of the robot. The goal of the path planning problem is to create a feasible and safe path, which is achieved by either minimizing or maximizing the cost.The goal of this work is to minimize the cost function.

In the proposed approach, since there is a randomness in the algorithm, there might be a case that the generated path doesn't reach the target. Therefore the cost function is chosen as follows:

\begin{itemize}
    \item $C_{length}$ is the cost incurred due to travelling from source to destination. The shorter the path, the less is the cost and thus more feasible it is for the robot to take that path.
    
    \item $C_{turn}$: It is the cost due to sharp turns during the journey. The trajectory has to be smooth since sharp turns leads to more fuel consumption, thereby increasing the energy input of the robot. 
    
    \item $C_{left}$ is the cost incurred due to not reaching the destination due to getting stopped by a dominant obstacle. This is the most important cost and it should be made zero for achieving the best cost.
\end{itemize}

Thus the total cost function can be shown as follows:\\
\small
{
\begin{eqnarray}
C_{Total} = k_{1} \times C_{length} + k_{2} \times C_{turns} + k_{3} \times C_{left}
\end{eqnarray}
}

where $k_1$, $k_2$, $k_3$ are constants.

\subsubsection{Assumption}
Some assumptions had been made in the research to get more efficient and useful results. The assumptions made are as follows:

\begin{itemize}
    \item The robot is considered as a point object.
    
    \item The speed of the robot is kept constant during the course of entire simulation. The robot moves faster than compared to targers and obstacles.
    
    \item Upon reaching the target point, a robot can stop immediately irrespective of the momentum of the robot.
\end{itemize}

\section{Results}

\subsection{Experiment design}

\subsubsection{Environment}
For starting the path planning problem, an environment needs to be set-up/simulated in which the robot can move. There are several properties associated with the environment like obstacles, target, number of robots etc. We have performed simulation using different environment in order to gain more clarity on the performance of the algorithm. There are two types of static obstacles, one that are created in advance and other that is created during run time. These obstacles mimic the real life scenario and can be though of as a building, tree, birds, airplane etc. For simplicity purpose the shape of the obstacles are taken to be cuboidal. The dynamic obstacles are point objects and move randomly in the search space. The robot is assumed to be a point object which can fly in 3D space. There can be multiple goals of different nature like stationary, moving and time varying. The 3D space is considered to be a grid, where each element of the grid is occupied if it contains an obstacle and free otherwise.

\begin{table}[h!]
\begin{center}
\caption{Default testing parameters}
\begin{tabular}{ | p{4cm} | p{2.5cm}|} 
\hline 
{\bf \begin{center} Parameter \end{center}} & 
{\bf  \begin{center} Value \end{center}}\\
\hline

Map Size & $64 \times 64 \times 64$\\
\hline
Maximum Path Finding Time & $\infty$\\
\hline
Max percent of static obstacles & 8 percent\\
\hline
Total targets & $3$\\
\hline

\end{tabular}
\end{center}
\end{table}

\begin{figure}[hbt!]
	\begin{center}
		\includegraphics[scale=0.99]{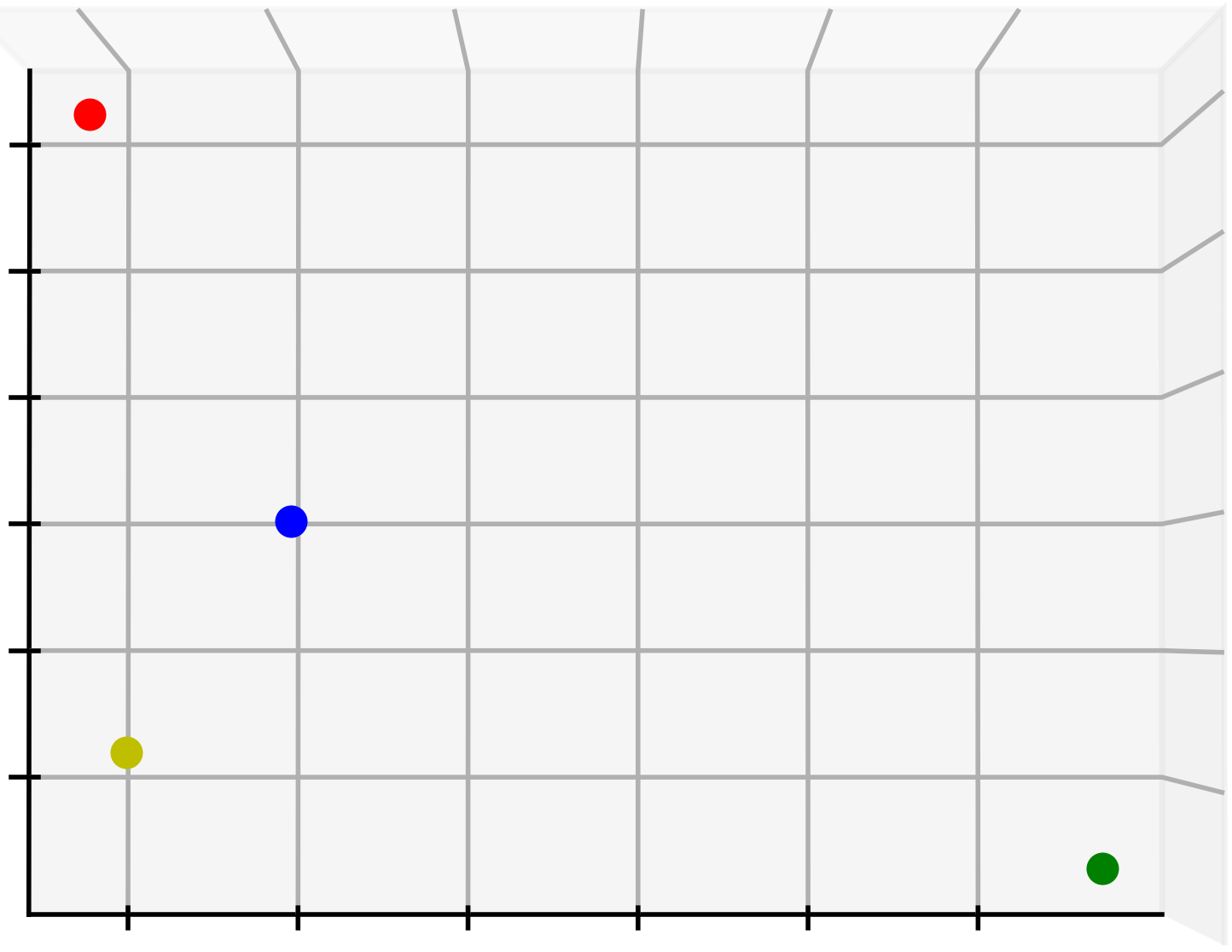}
		\caption{Environment having no obstacles}
	\end{center}
\end{figure}

\begin{figure}[hbt!]
	\begin{center}
		\includegraphics[scale=0.99]{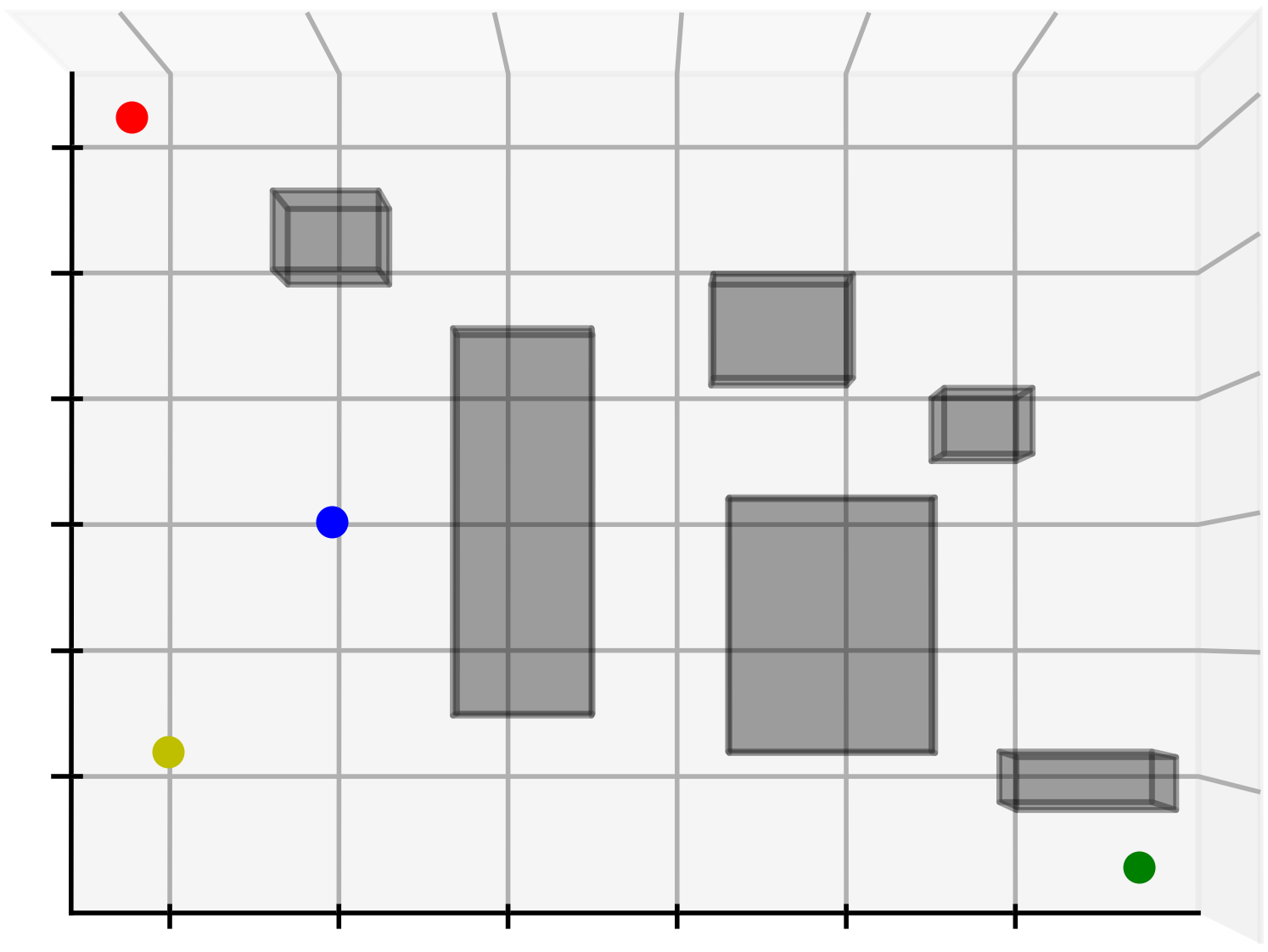}
		\caption{Environment having Fixed Static Obstacles(Cuboid)}
	\end{center}
\end{figure}

\begin{figure}[hbt!]
	\begin{center}
		\includegraphics[scale=0.99]{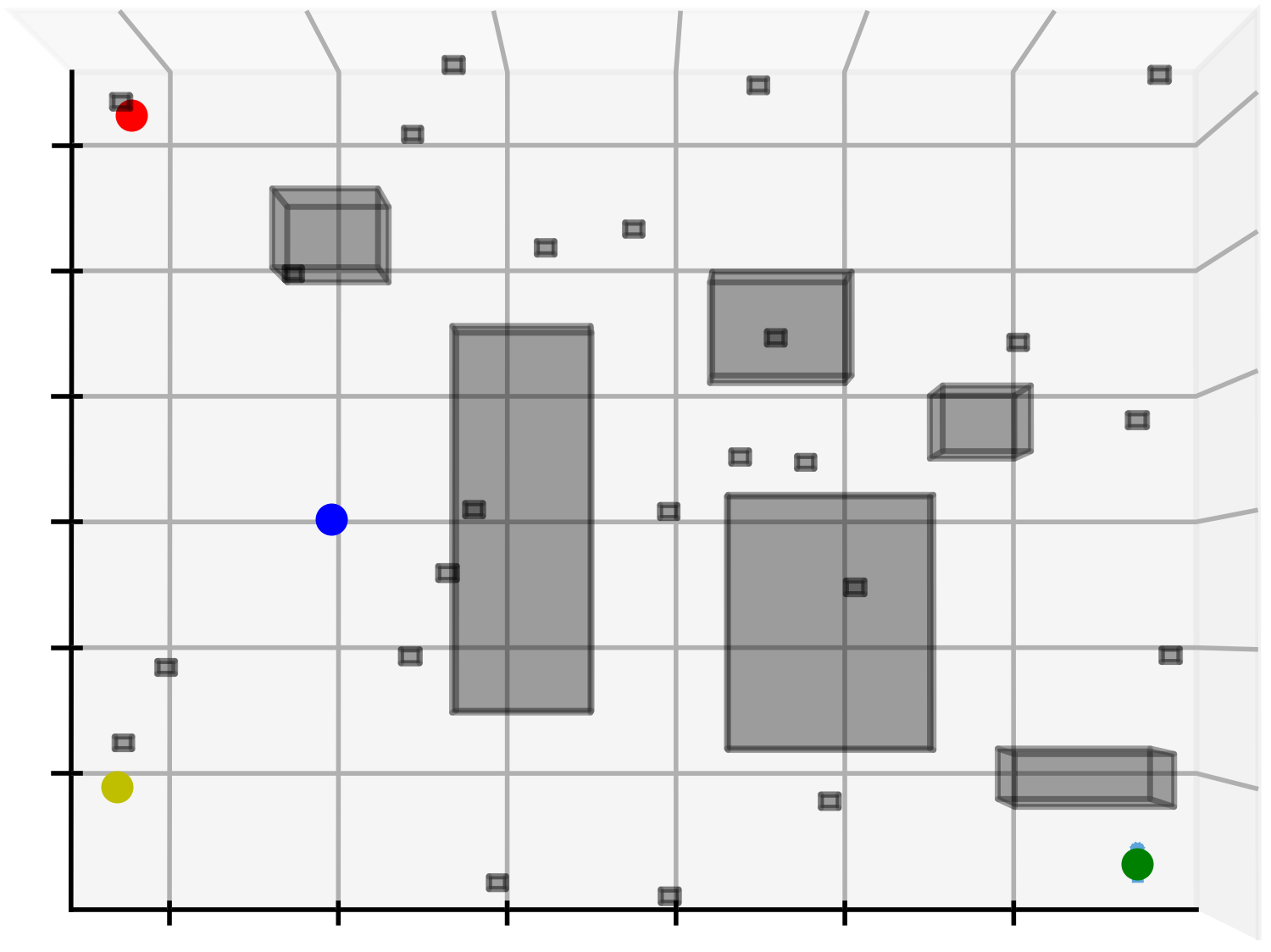}
		\caption{Environment having Fixed and Runtime Static(small cuboid) Obstacles}
	\end{center}
\end{figure}

\begin{figure}[hbt!]
	\begin{center}
		\includegraphics[scale=0.99]{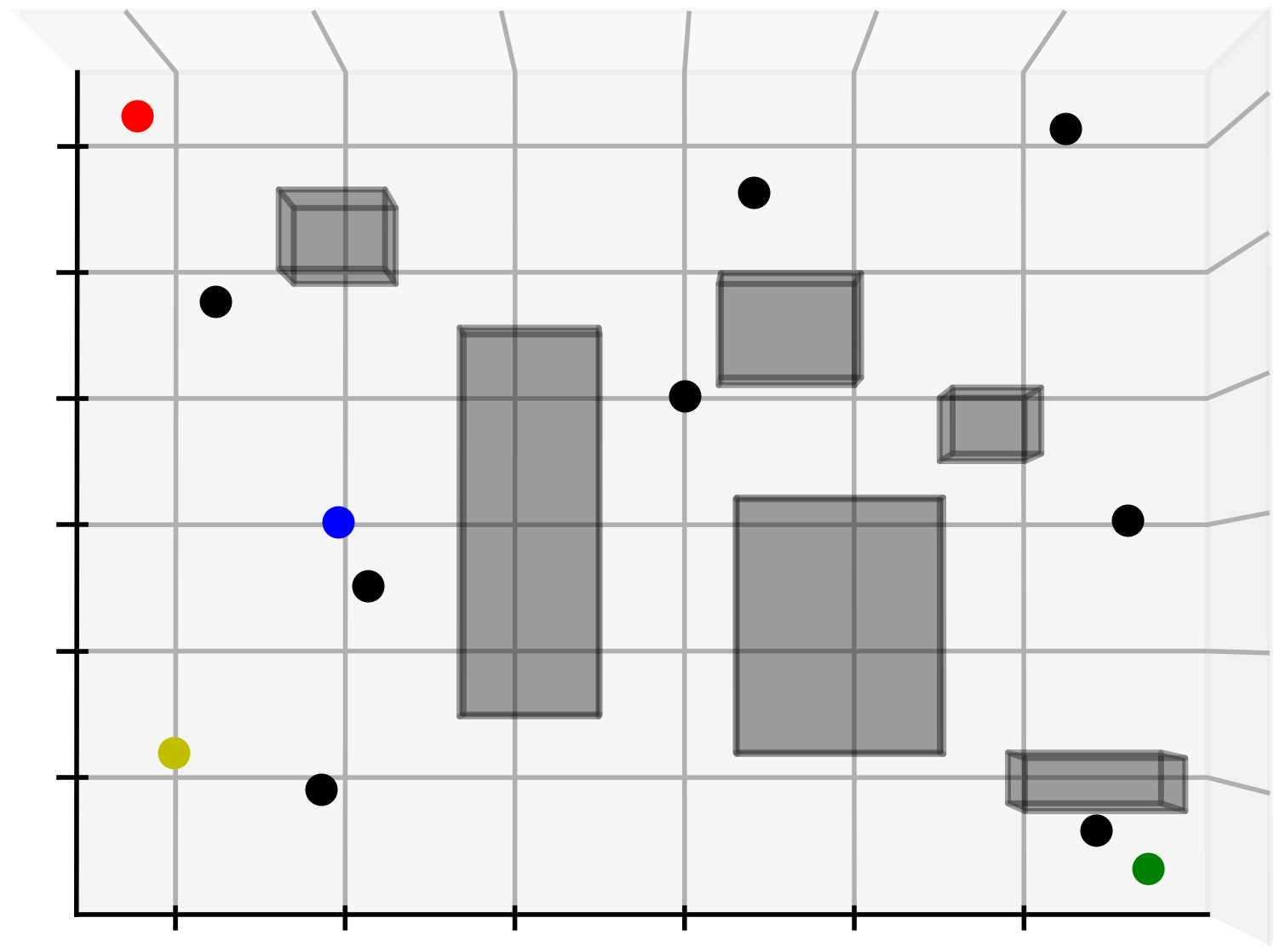}
		\caption{Environment having Dynamic Obstacles(black dot)}
	\end{center}
\end{figure}

\begin{figure}[hbt!]
	\begin{center}
		\includegraphics[scale=0.99]{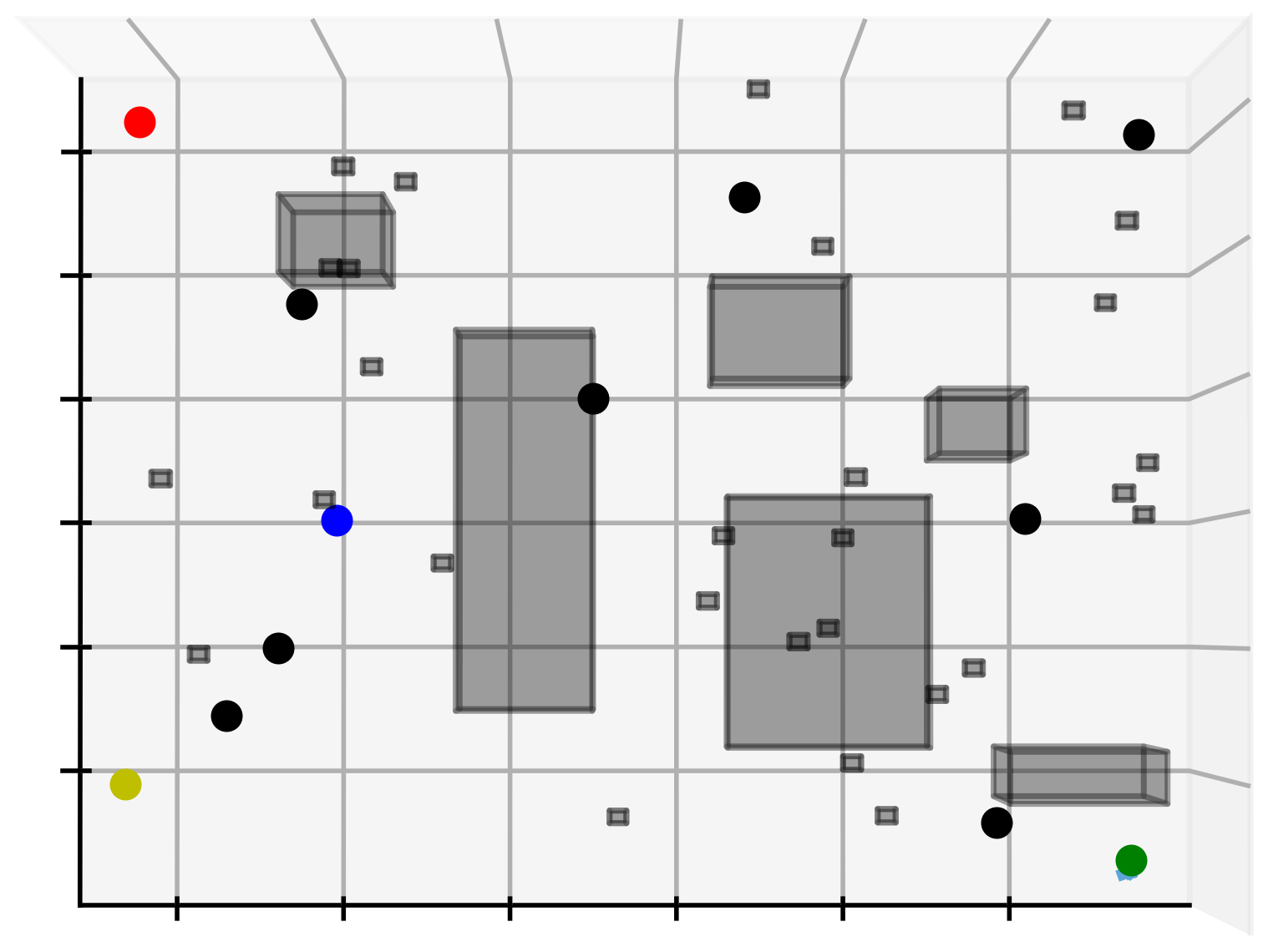}
		\caption{Environment having Dynamic and Runtime Static Obstacle}
	\end{center}
\end{figure}

Following represents the different components in the map:
\begin{itemize}
        \item Green Dot: Robot.
        \item Blue Dots: Time varying Goal.
        \item Red Dots: Static Goal.
        \item Yellow Dots: Dynamic Goal.
        \item Black Dots: Dynamic Obstacles.
        \item Grey Cuboid: Static Obstacles(FIO).
        \item Grey small Cube: Runtime Static Obstacles(FRO).
    \end{itemize}

\subsubsection{Experiment description}
The simulation has been performed in Python 2.7.9. All the simulations were done on computer with i7-2600 CPU @ 3.40GHz with 4GB RAM. The constants for the simulation are given in Table 1. 

\subsection{Experiment 1}
The first experiment consist of all the algorithms namely GSO, hGSO\footnotemark, IWO, hIWO\footnotemark[\value{footnote}], BBO and hBBO\footnotemark[\value{footnote}] are run in five different environments and cost incurred during path planning is noted. The path followed by the agent is shown in Figures 4-8 and the results can be seen in Tables 2 to 6. Table 2 shows the results for Map 1, which does not consist of any obstacle i.e. all nodes are free nodes. Table 3 shows the results for Map 2, which includes only one type of static obstacles known as FIO(Fixed Individual Obstacles). The position and dimensions of these obstacles are fixed. Table 4 tabulates the results for Map 3, which consists of static obstacles of two types, FIO, and FRO(Fixed Random Obstacles). FRO are generated at runtime and does not have any fixed positions and dimensions. The maximum percentage of nodes covered FRO that can be generated is limited to 8\%, as given in Table 1. Table 5 shows the results for Map 4, which consists of both dynamic and static obstacles(only FRO). Dynamic obstacles for the purposes of the simulation are considered point objects, are independent of each other, and their speeds are always less than the speed of the agent. Finally, Table 6 shows the results on Map 5, which contains dynamic obstacles as well as static obstacles of both the kinds(FIO and FRO). 

\footnotetext{here prefix \textit{h} stands for use of hierarchical levels}

%% TABLE 2
\noindent
\begin{table}[hbt!]
\caption{Simulation results for Map 1}
\begin{center}
\resizebox{\columnwidth}{!}{
\begin{tabular}{|l|*{3}{c|}}\hline
\backslashbox{\textbf{Algorithm}}{\textbf{Metric}}
&\makebox[8em]{\textbf{Elapsed  time(in  sec)}}&\makebox[7em]{\textbf{Expanded  Nodes}}&\makebox[3em]{\textbf{Cost}}\\ \hline

    GSO  & 11683 & 18488   & 4928032\\ \hline
    hGSO & 8811  & 25215   & 4482761\\ \hline
    IWO  & 32945 & 77200   & 11127117\\ \hline
    hIWO & 79798 & 122954  & 10503436\\ \hline
    BBO  & 20574 & 64268   & 4176324\\ \hline
    hBBO & 27401 & 82045   & 6296116\\ \hline

\end{tabular}
}
\label{table_2}
\end{center}
\end{table}

%% TABLE 3
\noindent
\begin{table}[hbt!]
\caption{Simulation results for Map 2}
\begin{center}
\resizebox{\columnwidth}{!}{
\begin{tabular}{|l|*{3}{c|}}\hline
\backslashbox{\textbf{Algorithm}}{\textbf{Metric}}
&\makebox[8em]{\textbf{Elapsed  time(in  sec)}}&\makebox[7em]{\textbf{Expanded  Nodes}}&\makebox[3em]{\textbf{Cost}}\\ \hline

    GSO  & 9957 & 19455   & 5034259\\ \hline
    hGSO & 10737  & 26568   & 4532053\\ \hline
    IWO  & 65392 & 97607   & 11442094\\ \hline
    hIWO & 65931 & 120754  & 8671267\\ \hline
    BBO  & 22611 & 78290   & 4335125\\ \hline
    hBBO & 32329 & 83000   & 4290042\\ \hline

\end{tabular}
}
\label{table_3}
\end{center}
\end{table}

%% TABLE 4
\noindent
\begin{table}[hbt!]
\caption{Simulation results for Map 3}
\begin{center}
\resizebox{\columnwidth}{!}{
\begin{tabular}{|l|*{3}{c|}}\hline
\backslashbox{\textbf{Algorithm}}{\textbf{Metric}}
&\makebox[8em]{\textbf{Elapsed  time(in  sec)}}&\makebox[7em]{\textbf{Expanded  Nodes}}&\makebox[3em]{\textbf{Cost}}\\ \hline

    GSO  & 10710 & 25729   & 4978856\\ \hline
    hGSO & 11141  & 26700   & 5024262\\ \hline
    IWO  & 60908 & 102656   & 11970281\\ \hline
    hIWO & 85845 & 128584  & 10390973\\ \hline
    BBO  & 26511 & 82502   & 5256216\\ \hline
    hBBO & 34931 & 92818   & 4766282 \\ \hline

\end{tabular}
}
\label{table_5}
\end{center}
\end{table}

%% TABLE 5
\noindent
\begin{table}[hbt!]
\caption{Simulation results for Map 4}
\begin{center}
\resizebox{\columnwidth}{!}{
\begin{tabular}{|l|*{3}{c|}}\hline
\backslashbox{\textbf{Algorithm}}{\textbf{Metric}}
&\makebox[8em]{\textbf{Elapsed  time(in  sec)}}&\makebox[7em]{\textbf{Expanded  Nodes}}&\makebox[3em]{\textbf{Cost}}\\ \hline

    GSO  & 13703 & 27837   & 4850499\\ \hline
    hGSO & 10094  & 25279   & 4507687\\ \hline
    IWO  & 65050 & 95305   & 8465725\\ \hline
    hIWO & 85286 & 130057  & 11139622\\ \hline
    BBO  & 21972 & 82746   & 3952487\\ \hline
    hBBO & 37490 & 98404   & 7572836\\ \hline

\end{tabular}
}
\label{table_5}
\end{center}
\end{table}

%% TABLE 6
\noindent
\begin{table}[hbt!]
\caption{Simulation results for Map 5}
\begin{center}
\resizebox{\columnwidth}{!}{
\begin{tabular}{|l|*{3}{c|}}\hline
\backslashbox{\textbf{Algorithm}}{\textbf{Metric}}
&\makebox[8em]{\textbf{Elapsed  time(in  sec)}}&\makebox[7em]{\textbf{Expanded  Nodes}}&\makebox[3em]{\textbf{Cost}}\\ \hline

    GSO  & 11880 & 24200   & 5164557\\ \hline
    hGSO & 10847  & 24465   & 4992658\\ \hline
    IWO  & 78163 & 111094   & 12570570\\ \hline
    hIWO & 84452 & 116774  & 10294457\\ \hline
    BBO  & 41883 & 100409   & 5364178\\ \hline
    hBBO & 32184 & 86921   & 4299352\\ \hline

\end{tabular}
}
\label{table_6}
\end{center}
\end{table}

\subsection{Experiment 2}
The second set of experiments are performed on Map 5, which has dynamically, and obstacles generated randomly at runtime. The results of different algorithms are compared in Table 7-9. The experiments are performed on Map 5 for 30\%, 25\%, 20\% obstacle density for Table 7-9, respectively.

%% TABLE 7
\noindent
\begin{table}[hbt!]
\caption{30\% Obstacle Density}
\begin{center}
\resizebox{\columnwidth}{!}{
\begin{tabular}{|l|*{3}{c|}}\hline
\backslashbox{\textbf{Algorithm}}{\textbf{Metric}}
&\makebox[8em]{\textbf{Elapsed  time(in  sec)}}&\makebox[7em]{\textbf{Expanded  Nodes}}&\makebox[3em]{\textbf{Cost}}\\ \hline

    GSO  & 7132 & 23423   & 4743890\\ \hline
    hGSO & 10522  & 31077   & 4564806\\ \hline
    IWO  & 45957 & 130538   & 13190259\\ \hline
    hIWO & 43780 & 123058  & 11576077\\ \hline
    BBO  & 28994 & 88954   & 5289164\\ \hline
    hBBO & 40917 & 119277   & 5487862\\ \hline

\end{tabular}
}
\label{table_7}
\end{center}
\end{table}

%% TABLE 8
\noindent
\begin{table}[hbt!]
\caption{25\% Obstacle Density}
\begin{center}
\resizebox{\columnwidth}{!}{
\begin{tabular}{|l|*{3}{c|}}\hline
\backslashbox{\textbf{Algorithm}}{\textbf{Metric}}
&\makebox[8em]{\textbf{Elapsed  time(in  sec)}}&\makebox[7em]{\textbf{Expanded  Nodes}}&\makebox[3em]{\textbf{Cost}}\\ \hline

    GSO  & 1632 & 24013   & 4955365\\ \hline
    hGSO & 3048  & 33368   & 5267612 \\ \hline
    IWO  & 15268 & 82252   & 6551911\\ \hline
    hIWO & 24451 & 120535  & 10661109\\ \hline
    BBO  & 14793 & 106620   & 5255228\\ \hline
    hBBO & 28616 & 123523   & 5183095\\ \hline

\end{tabular}
}
\label{table_8}
\end{center}
\end{table}

%% TABLE 9
\noindent
\begin{table}[hbt!]
\caption{20\% Obstacle Density}
\begin{center}
\resizebox{\columnwidth}{!}{
\begin{tabular}{|l|*{3}{c|}}\hline
\backslashbox{\textbf{Algorithm}}{\textbf{Metric}}
&\makebox[8em]{\textbf{Elapsed  time(in  sec)}}&\makebox[7em]{\textbf{Expanded  Nodes}}&\makebox[3em]{\textbf{Cost}}\\ \hline

   GSO  & 2714 &  25692  & 5481396\\ \hline
    hGSO & 3094  & 29198   & 5034207\\ \hline
    IWO  & 17813 & 110994   & 10362583\\ \hline
    hIWO & 23787 & 122169  & 8515086\\ \hline
    BBO  & 58434 & 97418   & 6092502 \\ \hline
    hBBO & 23921 & 100271   & 5963980\\ \hline

\end{tabular}
}
\label{table_9}
\end{center}
\end{table}

\section{Conclusion and future scope}
We hereby conclude that we have successfully developed a system that predicts the optimised path based on the environment. The hierarchical version of GSO has proved to give better cost results, in most of the cases, when compared to the normal GSO. hGSO is also better when compared to IWO and BBO and there respective hierarchical versions. The research can be extended to different swarm optimization algorithms communicating with each other and searching the target. Other NIA can also be applied in order to solve the problem of path planning. The multi robot coordination can be enhanced by diving the environment into different regions, with each region having different number of robots which interact with robots in same cluster as well as robots outside the cluster.

%%%%%%%%%%%%%%%%%%%
\bibliographystyle{ACM-Reference-Format}
\bibliography{bibliography}
%%%%%%%%%%%%%%%%%%%

\end{document}